# Leveraging Multimodal Self-Consistency Reasoning in Coding Motivational Interviewing for Alcohol Use Reduction

Guangzeng Han, MS[1], ghan@memphis.edu

James G. Murphy, PhD[4], jgmurphy@memphis.edu

Benjamin O. Ladd, PhD[5], benjamin.ladd@wsu.edu

Xiaolei Huang, PhD[1], xiaolei.huang@memphis.edu

Brian Borsari, PhD[2,3], brian.borsari@va.gov

**Affiliations:**

[1]Department of Computer Science, University of Memphis, 3725 Norriswood Ave, Memphis, TN, 38152

[2]Veterans Affairs Health Care System, 4150 Clement Street, San Francisco, CA 94121

[3]Department of Psychiatry and Behavioral Sciences, University of California San Francisco, 675 18th Street, San Francisco, CA 94143.

[4]Department of Psychology, University of Memphis, Memphis, 400 Fogelman Drive, TN, 38152

[5]Department of Psychology, Washington State University Vancouver, 14204 NE Salmon Creek Ave, Vancouver, WA, 98686

## Abstract

**Background:** Understanding client behaviors and predicting their outcomes require coding Motivational interviewing (MI) sessions with intensive labor costs and time consumptions of MI professionals. The advances in audio language models (ALMs) open promising opportunities in automating coding process, capturing critical multimodal signals of behavioral patterns, and interactive collaborations with expert annotators.

**Objective:** This study aims to develop an automatic motivational interviewing (MI) coding approach based on ALMs that analyzes raw audio input and integrates predictions from multiple reasoning trajectories using self-consistency to improve coding robustness.

**Methods:** We experimented with 5 recorded sessions from de-identified MI audio tapes. We deployed audio-language models with four complementary analytic prompts to augment utterance-level reasoning: *analytic* (verbal cues), *prosody-aware* (acoustic cues), *evidence-scoring* (quantitative hypothesis test), and *comparative* (contrastive reasoning). Three stochastic samples were drawn per prompt, generating 12 independent reasoning trajectories per utterance, and majority votes of all trajectories decided predictions.

**Results:** Performance was evaluated using accuracy, precision, recall, and macro-F1 scores. The proposed MM-SC approach achieved 52.56% accuracy, 54.03% precision, 47.45% recall, and a macro-F1 score of 46.40%, exceeding the performance of baseline methods. Systematic ablation removing individual modules consistently degraded performance on primary metrics.

**Conclusions:** Multimodal self-consistency outperforms single-pass baseline prompting approaches. These findings suggest that incorporating both what clients say and how they say it can support more reliable MI coding.

## 1. Introduction

Motivational interviewing (MI)[1–3] is a collaborative, client-centered counseling approach designed to strengthen personal motivation and commitment to behavioral change. Originally developed for addiction treatment, MI is now widely used across clinical psychology[4], public health, and behavioral medicine.[5] A core element of MI is the counselor's ability to recognize and respond to clients' language about change.[6] In this context, an utterance refers to a contiguous segment of client speech expressing a single communicative intent, typically bounded by pauses or speaker turns. Client utterances that support change are coded as Change Talk (CT), those that argue against change are coded as Sustain Talk (ST), and utterances without clear motivational direction are coded as Follow/Neutral (FN).[3] While prior automatic MI coding studies[7,8] leveraged linguistic and acoustic modalities, these methods typically operate as static feature fusion[9,10] pipelines that combine linguistic and acoustic information in a single-pass prediction process. This forces the model to make an immediate decision without resolving conflicting evidence; for instance, if a client says, "I suppose I could try" (textually positive) but speaks with a flat, hesitant tone (acoustically negative), a standard fusion model often struggles to reconcile these clashing signals.

Recent advances in large language models (LLMs)[11,12] and audio-language models[13] (ALMs) offer new opportunities for automatic MI coding. However, direct application of such models to MI remains challenging. Models often lack mechanisms to integrate analytic reasoning with complementary sources of evidence, such as prosodic affect (e.g., intonation, pauses and emphasis), as well as comparative evaluation of MI codes, all of which human coders naturally use. To address these limitations, we draw inspiration from the self-consistency method[14], originally proposed to improve the reliability of language model reasoning by sampling multiple reasoning paths under the same prompt and selecting the most consistent outcome across them. We extend this idea to multimodal MI coding by enforcing consistency both across repeated samples from the same prompt and across multiple, clinically motivated prompts that analyze the same raw audio from complementary perspectives. These perspectives are designed to reflect complementary strategies used by human MI coders, allowing the model to examine each utterance from multiple angles before arriving at a final code.

Our method processes raw audio directly without generating transcripts. We evaluate the approach on five MI counseling sessions drawn from recorded audio tapes.[1] The contributions of this work are threefold: 1) we introduce a multimodal self-consistency method (MM-SC) for automatic MI coding that integrates complementary reasoning perspectives; 2) we demonstrate that prompt-level diversity improves stability and robustness on spontaneous counseling speech; and 3) we conduct systematic ablation analyses showing that each reasoning pathway contributes uniquely to overall performance.

## 2. Data

The data used in this study are drawn from previously collected MI audio recordings of counseling sessions between trained counselors and college students discussing alcohol use reduction or cessation.[1] Each session includes a full-session audio recording and a verbatim transcript that was manually produced and annotated by trained human coders using the Motivational Interviewing Skill Code, providing utterance-level ground-truth labels (CT, ST, or FN).[14]

The original transcripts do not include precise utterance-level time boundaries in the audio. To align the human-coded utterances with the corresponding audio segments, we applied the OpenAI Whisper model[15] to obtain word-level timestamps and then aligned the automatically generated transcripts with the original human transcripts using fuzzy string matching to correct minor recognition and formatting differences. This alignment procedure allowed us to segment the continuous session audio into accurate utterance-level audio clips without performing speaker diarization. The sessions analyzed in this study represent a quality-controlled subset of the available corpus, selected to ensure reliable utterance–audio alignment; sessions with insufficient temporal correspondence between transcripts and audio were excluded, as accurate alignment is required for evaluating the proposed multimodal method. The resulting dataset consists of audio segments paired with human-annotated MI codes for each client utterance, with summary statistics reported in Table 1.

Table 1: Summary of dataset statistics.

| Session (N) | Words (Avg per Utterance, N) | Audio (Avg per Utterance, seconds) | Change Talk (Total Utterances, N) | Sustain Talk (Total Utterances, N) | Follow/Neutral (Total Utterances, N) |
|---|---|---|---|---|---|
| 5 | 31.5 | 7.3 | 371 | 135 | 392 |

## 3. Methods

This section describes the proposed multimodal self-consistency (MM-SC) framework for utterance-level motivational interviewing (MI) coding and the experimental setup used for evaluation. We first introduce the overall architecture of MM-SC and its two core modules. We then describe the prompting strategies that guide multi-perspective reasoning over raw audio input, followed by the aggregation mechanism that produces the final prediction via self-consistency. Finally, we present the experimental setup, baseline methods, and evaluation metrics used to assess model performance.

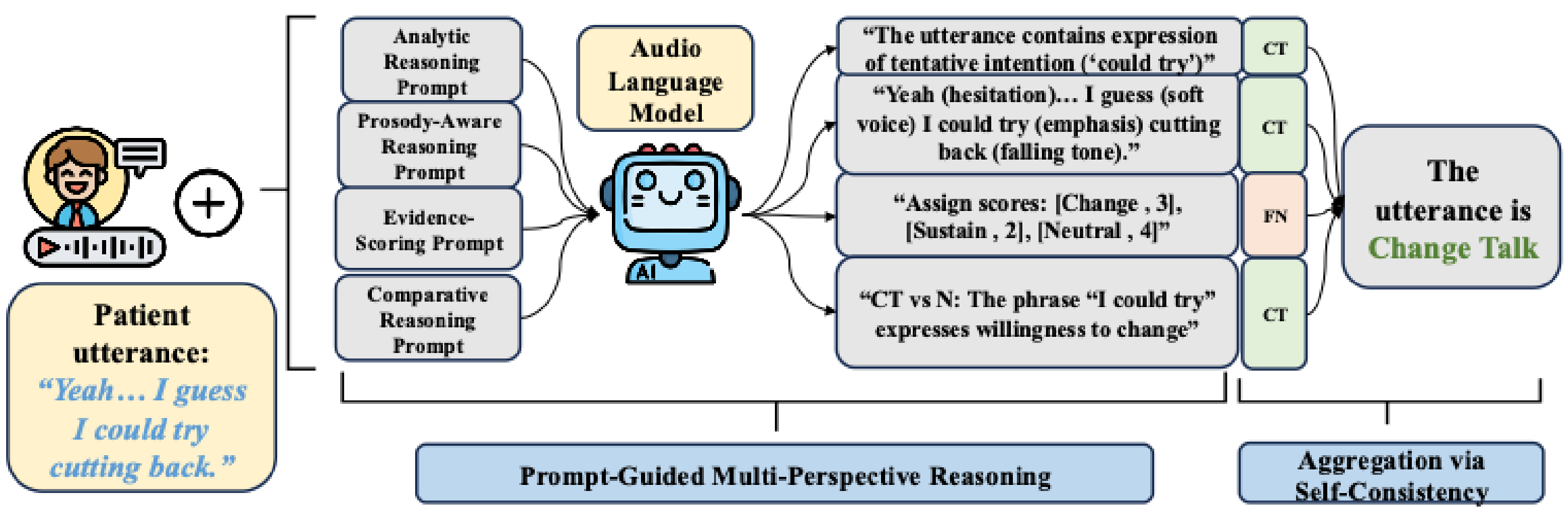

Figure 1: Overview of the proposed method, which applies four prompting strategies to audio language model and aggregates predictions through self-consistency.

## 3.1 Overview of MM-SC

We address utterance-level motivational interviewing (MI) coding by developing an end-to-end multimodal method that predicts CT, ST, or FN directly from raw audio input. Figure 1 shows our proposed MM-SC method operates through two integrated modules. The first module, *Prompt-guided Multi-perspective Reasoning*, employs four distinct prompting strategies to query the ALM. This process analyzes the raw audio through complementary lenses to generate independent reasoning chains focusing on semantic content, prosodic delivery, and comparative evidence. The second module, *Aggregation via Self-Consistency*, aggregates these diverse predictions via majority voting to filter out stochastic errors and capture the true motivational intent that remains invariant across different analytical viewpoints.

## 3.2 Prompt-Guided Multi-Perspective Reasoning

Client intentions in MI can emerge from the interaction between what a client says and how the utterance is delivered.[7,16,17] To capture both sources of information and to handle situations in which verbal and acoustic cues diverge, our method queries the ALM using four complementary prompting strategies. Each prompt guides the model to analyze the same raw audio segment through a different reasoning pathway, encouraging it to consider semantic content, prosodic delivery, or comparative evidence from distinct perspectives. Although these pathways differ in focus, a reliable motivational interpretation should lead the ALM to converge on the same MI code across them. This cross-prompt agreement is the form of consistency our method seeks to measure and later aggregate. The full prompt texts are in Appendix Table S1.

*Analytic Reasoning Prompt* **(P1)** encourages the model to focus on the literal content of the utterance. It guides the ALM to identify expressions commonly associated with motivational direction in MI, such as statements showing desire, ability, reasons, need, commitment, or indications of steps toward change. Based on these cues, the model determines whether the client is moving toward change (CT), defending the status quo (ST), or providing information that does not reflect a clear motivational stance (FN). *Prosody-Aware Prompt* **(P2)** extends this type of analysis by drawing attention to vocal qualities such as hesitation, pauses, and intonation. Rather than specifying these cues in detail, the prompt reminds the model that vocal tone can reinforce or contradict word literal indications. This allows the ALM to incorporate acoustic evidence when semantic and prosodic signals point in different directions. The remaining two prompts introduce a more structured comparative perspective. *Evidence-Scoring Prompt* **(P3)** requires the ALM to reason through the utterance and assign an independent strength score to CT, ST, and FN before selecting the most possible code. This process encourages the model to explicitly evaluate competing interpretations rather than settling on the first plausible option.

*Comparative Reasoning Prompt* **(P4)** asks the model to contrast the three categories directly, examining whether the utterance indicates movement toward change, toward maintaining current behavior, or toward neither. This approach is intended to reduce reliance on the FN category in situations where the evidence is mixed or partially conflicting.

Our prompting strategies guide the ALM to reinterpret each utterance through multiple clinically motivated perspectives with semantics, vocal delivery, and graded evidence. The resulting set of diverse predictions provides the foundation for the Multimodal Self-Consistency module in Section 3.2, which aggregates perspectives to identify behavioral patterns that remain consistent across analytical viewpoints.

### 3.3 Aggregation via Self-Consistency

When the ALM analyzes an utterance, it may give slightly different outputs across runs due to randomness in the model generation process. Each of the four prompts in Section 3.1 examines a different aspect of the same audio segment. Thus, a single prediction may not necessarily reflect the overall pattern, while our model will reconsider the same utterance from multiple angles. To obtain a more dependable decision, we examine whether the ALM arrives at the same label across these different reasoning paths.

For each utterance, we run the ALM under all four prompts and generate three responses per prompt, yielding twelve predictions in total. Each prediction represents one possible way the model interprets the audio. Some predictions may focus more on wording, others on tone or the strength of evidence. The final label is obtained by the majority votes that counts CT, ST, and FN appear among the twelve predictions and assign the label with the highest count. This procedure identifies the conclusion that the ALM repeats most consistently across prompts and repeated runs, and it reduces the influence of any single prompt or sampling variation. When two labels receive the same number of votes, we break the tie by choosing CT or ST over FN, so that the system does not default to FN when there are repeated signs of movement toward or away from change.

### 3.4 Experimental Setup and Evaluation

All experiments use the Qwen3-Omni-30B-A3B-Instruct model.[13] Each utterance is provided to the model as a standalone raw audio segment. For every prompt in our method, we generate three predictions using fixed hyperparameters for the ALM, the temperature is set to 1.0 and top-p is 0.5, which control the randomness of text generation. We compare MM-SC against two baseline prompting strategies: Direct Prompt and Chain-of-Thought (COT) Prompt.[18] Direct Prompt feeds the ALM with raw audio and a single instruction to predict CT, ST, or FN. Chain-of-Thought (COT) breaks a complex task into multiple easier sub-tasks. We report accuracy, precision, recall, and macro-F1 scores, with accuracy and macro-F1 serving as the primary metrics.

## 4. Results

### 4.1 Main Results

Table 2: Main results for utterance-level motivational interviewing coding. [1]

| METHOD | ACCURACY | PRECISION | RECALL | MACRO-F1 |
|---|---|---|---|---|
| DIRECT (BASELINE) | 46.22 | 47.22 | 45.05 | 42.48 |
| COT (BASELINE) | 49.89 | 50.32 | 44.03 | 41.97 |
| MM-SC (EXPERIMENTAL) | 52.56 | 54.03 | 47.45 | 46.40 |

[1] Accuracy and Macro-F1 score are the primary metrics. All metrics are reported as percentages. DIRECT denotes the direct prompting baseline; COT denotes chain-of-thought prompting; MM-SC denotes the proposed multimodal self-consistency approach. Prompt templates are provided in Table S1.

Table 2 reports the main results for utterance-level motivational interviewing coding, comparing the proposed multimodal self-consistency method (MM-SC) with two baseline prompting strategies: Direct prompting and chain-of-thought (COT) prompting. All metrics are reported as percentages, and the total number of utterances by MI code is provided in Table 1. MM-SC achieves the highest performance across all 4 metrics. In particular, MM-SC attains a macro-F1 score of 46.40%, compared with 42.48% for the Direct baseline and 41.97% for the COT baseline. Accuracy under MM-SC reaches 52.56%, exceeding that of Direct prompting (46.22%) and COT prompting (49.89%). Precision and recall also show consistent improvements under MM-SC relative to both baseline methods.

### 4.2 Ablation Study

Table 3: Ablation results of the proposed multimodal self-consistency approach.[2]

| METHOD | ACCURACY | PRECISION | RECALL | MACRO-F1 |
|---|---|---|---|---|
| **MM-SC (Audio + Text)** | 54.68 | 54.23 | 50.37 | 50.17 |
| **MM-SC (Audio)** | 52.56 | 54.03 | 47.45 | 46.40 |
| **MM-SC (Text)** | 50.67 | 54.32 | 46.49 | 43.53 |
| **w/o P1** | 51.45 | 56.81 | 44.84 | 42.63 |
| **w/o P2** | 50.78 | 51.43 | 46.06 | 44.20 |
| **w/o P3** | 50.11 | 56.68 | 43.45 | 40.76 |
| **w/o P4** | 50.78 | 50.59 | 46.09 | 44.61 |

An ablation study is conducted to examine the effects of prompting strategies and input modality within the MM-SC framework. For modality ablations, we compare the full MM-

[2] All metrics are reported as percentages. MM-SC denotes the full multimodal self-consistency method. MM-SC (Audio + Text) uses both of audio and transcript input. MM-SC (Audio) uses audio input only. MM-SC (Text) uses transcript input instead of audio and excludes the prosody-aware prompt (P2). w/o P1–P4 denote ablation variants that remove Prompt 1–Prompt 4, respectively, from the full MM-SC framework.

SC system operating on raw audio only (MM-SC (Audio)) with a text-only variant that replaces audio input with transcripts (MM-SC (Text)), in which the prosody-aware prompt (P2) is omitted. We further evaluate an augmented setting, MM-SC (Audio + Text), in which both raw audio and transcripts are provided as input. For prompting ablations, each reasoning strategy (w/o P1, w/o P2, w/o P3, w/o P4) is removed one at a time while the remaining prompts and audio input are held constant. Table 3 summarizes the performance of all configurations. Among all evaluated variants, MM-SC (Audio + Text) achieves the highest overall performance, with an accuracy of 54.68% and a macro-F1 score of 50.17%. The audio-only MM-SC system attains an accuracy of 52.56% and a macro-F1 score of 46.40%. Replacing audio input with transcripts results in lower overall performance, with MM-SC (Text) achieving a macro-F1 score of 43.53%. All prompt removal variants exhibit reduced macro-F1 scores relative to the full MM-SC (Audio) system, with macro-F1 values ranging from 40.76% (w/o P3) to 44.61% (w/o P4). Precision and recall vary across ablation settings; however, none of the ablated variants exceeds the full MM-SC systems in either accuracy or macro-F1.

### 4.3 Qualitative Error Analysis

We conducted a qualitative inspection of model predictions to identify common error patterns in utterance-level motivational interviewing (MI) coding. Two recurring patterns were observed: overthinking and underthinking. Overthinking occurs when reflective utterances containing evaluative language are classified as Change Talk despite lacking explicit motivational intent. For example, a client may state, “I guess things were better for me back in my freshman year.” Under MI coding guidelines, such statements are typically labeled as Follow/Neutral because they reflect evaluation of past experiences without indicating a desire or commitment to change. Underthinking occurs when subtle change-oriented implications related to drinking are classified as Follow/Neutral. For example, a client may state, “When I drink during the week, it really messes with my classes the next day.” Although no explicit commitment to change is expressed, human MI coders often label such statements as Change Talk because they describe negative consequences of drinking that are relevant to behavior change.

## 5. Discussion

### 5.1 Why MM-SC Outperforms Baseline Prompting Methods

The proposed multimodal self-consistency (MM-SC) approach consistently outperforms direct prompting and chain-of-thought (COT) prompting across all evaluation metrics. Direct prompting requires the model to assign an MI code based on a single pass over the audio input. In practice, many MI utterances are short, tentative, or linguistically underspecified, and their motivational direction depends on subtle cues such as hesitation, prosody, or partial commitment. Under direct prompting, the model must implicitly balance these cues within one unconstrained inference, making predictions sensitive to surface wording or transient acoustic signals. COT prompting introduces an explicit reasoning format, but the reasoning remains linear and tied to a single interpretive path. In MI coding, where an utterance may simultaneously exhibit weak signals of both change and sustain talk, a single chain of reasoning can prematurely commit to one interpretation without systematically considering alternatives. In contrast, MM-SC decomposes the coding task into multiple complementary reasoning perspectives that separately attend to semantic content, vocal delivery, and comparative evidence across MI codes. By aggregating predictions across these perspectives and repeated samples, MM-SC effectively performs cross-checking among alternative interpretations. This design reduces the impact of idiosyncratic reasoning paths and leads to more stable utterance-level classifications, which is reflected in its consistent improvements across accuracy, precision, recall, and macro-F1.

### 5.2 Insights from Ablation Analysis

The ablation analysis examines the effects of both input modality and prompting strategies within the MM-SC framework. Among all evaluated configurations, MM-SC (Audio + Text) achieves the highest accuracy and macro-F1 score, indicating that combining complementary linguistic and acoustic information yields the strongest overall performance when assessed using these primary metrics. MM-SC (Audio) outperforms the text-only configuration, MM-SC (Text), in both accuracy and macro-F1, demonstrating the importance of acoustic information for utterance-level MI coding. At the same time, MM-SC (Audio) also exceeds all reduced prompting variants on these primary metrics, indicating that the proposed framework remains robust even without access to transcripts.

Some ablated variants attain higher values on individual metrics such as precision or recall than MM-SC (Audio). However, these gains reflect trade-offs in labeling behavior rather than genuine improvements in overall coding performance. In utterance-level MI coding, increasing precision often corresponds to more conservative assignment of Change Talk, whereas increasing recall may reflect more liberal labeling. Such shifts can improve a single metric while simultaneously degrading performance on other classes. Accuracy and macro-F1 are therefore emphasized as primary evaluation metrics. Accuracy captures overall correctness across all utterances, while macro-F1 provides a balanced assessment across MI categories by jointly accounting for precision and recall under class imbalance.

The fact that none of the ablated variants exceeds MM-SC (Audio) on either accuracy or macro-F1 indicates that removing individual prompting strategies may sharpen specific decision tendencies, but at the cost of reduced balance and stability across classes. By contrast, the full MM-SC framework integrates complementary reasoning signals through aggregation, mitigating these trade-offs and enabling more robust utterance-level coding than any individual ablation setting.

### 5.3 Interpretation of Error Patterns

Motivational interviewing (MI) coding relies on detailed annotation guidelines that distinguish subtle differences in motivational intent. Without explicit knowledge of these coding rules, large language models or audio-language models may struggle to control their reasoning paths when interpreting client utterances. The overthinking errors suggest that the model may overinterpret evaluative language as evidence of motivational intent, while the underthinking errors indicate that it may fail to recognize implicit indications of change-related intent and instead treat statements describing consequences of drinking as purely informational. These patterns suggest that such errors may arise when the model lacks sufficient understanding of MI coding principles.

### 5.4 Limitations and Future Work

Several limitations should be considered when interpreting these findings. First, the evaluation was conducted on a small number of sessions drawn from a specific population, namely college students participating in alcohol-related motivational interviewing. Second, consistent with prior work[19,20] on automatic MI coding, performance at the utterance level remains constrained by the inherent difficulty of inferring motivational intent from short, context-dependent speech segments. Even for human coders, such utterances often require broader conversational context and clinical judgment to interpret reliably. Future work will explore two complementary directions. One direction focuses on improving practical performance by providing additional contextual information and incorporating more explicit motivational interviewing knowledge. The other direction will evaluate the proposed MM-SC framework in military and veteran settings, contingent on access to appropriate data, where motivational interviewing is widely used to address substance use, mental health concerns, and behavioral readiness.

## 6. Conclusion

This study presents a new approach for automatic motivational interviewing coding that analyzes patients' speech directly, without requiring manual transcripts. By combining multiple complementary analyses of the same utterance, the approach produces more consistent MI utterance categories and outperforms baseline methods across all experimental conditions. The results support the idea that reliable MI coding benefits from considering both what clients say and how they say it, and from comparing multiple perspectives rather than relying on a single decision. The proposed method offers a practical step toward automatic MI coding, particularly in settings where high-quality transcripts are unavailable. Future work will evaluate the method on larger MI corpora[2] and explore human-LLM collaborations for MI.

**Acknowledgments:** Dr. Borsari's contribution was supported by the San Francisco VA Health Care System. Mr. Han is supported by NSF (National Science Foundation) CNS-2318210, and the work is partially supported by NSF TI-2434589 and Dr. Huang's startup. The authors would thank for the organizers of Military Health System Research Symposium 2025 and their invite. The content is solely the responsibility of the authors and does not necessarily represent the official views of NSF, NIH, the Department of Veterans Affairs, Veterans Health Administration, Office of Academic Affiliations or the United States Government. We thank the computing resources provided by the iTiger GPU cluster[21] supported by the NSF MRI program under the award CNS-2318210.

**Funding:** National Science Foundation CNS-2318210 and TI-2434589

**Prior Presentation:** MHSRS ID: MHSRS-25-13819

# Appendix

Table S1: Prompt templates.[3]

| Prompt Name | Prompt Template |
|---|---|
| **Analytic Reasoning Prompt (Experimental Prompt 1)** | Task Definition: *[same]*.<br>Method: Think step by step about the meaning of the utterance and decide whether it reflects CT, ST, or FN. |
| **Prosody-Aware Prompt (Experimental Prompt 2)** | Task Definition: *[same]*.<br>Method: Paraphrase the utterance and annotate prosodic cues inline (e.g., hesitation, pause, rising tone), then reason about CT/ST/FN using both content and acoustic cues. |
| **Evidence-Scoring Prompt (Experimental Prompt 3)** | Task Definition: *[same]*.<br>Method: Provide step-by-step reasoning and assign independent 1–5 scores to CT, ST, and FN based on observed evidence; choose the highest. |
| **Comparative Reasoning Prompt (Experimental Prompt 4)** | Task Definition: *[same]*.<br>Method: Compare how well the utterance fits CT, ST, and FN by focusing on discriminative evidence; select the best-fitting label. |
| **Direct Prompt (Baseline)** | Task Definition: *[same for all prompts]*.<br>Method: Directly classify the utterance as CT, ST, or FN without providing reasoning. |
| **Chain-of-Thought Prompt (Baseline)** | Task Definition: *[same]*.<br>Method: Think step by step to decide CT, ST, or FN. |

[3] The Task Definition is detailed description of Change Talk (CT), Sustain Talk (ST) and Follow/Neutral (FN).